\documentclass[12pt]{article}

\usepackage{sbc-template}

\usepackage[english]{babel}
\usepackage[utf8]{inputenc}

\usepackage{graphicx}
\usepackage{url}
\usepackage{amsmath,amssymb,amsfonts}
\usepackage{booktabs}
\usepackage{array}
\usepackage{tabularx}
\usepackage{makecell}
\usepackage{siunitx}
\usepackage{balance}
\usepackage{subfigure}

\title{A Study of Action-Sequence Diffusion for Open-Loop Control under Dry Friction and Stiction}

\author{Eric Aislan Antonelo\inst{1}}
\address{
Department of Automation and Systems Engineering (EAS)\\
Federal University of Santa Catarina (UFSC)\\
Florianópolis, SC, Brazil\\
\email{eric.antonelo@ufsc.br}
}

\begin{document}

\maketitle

% \institute{Automation and Systems Engineering Department, Federal University of Santa Catarina, Florian\'opolis, Brazil\\
% \email{eric.antonelo@ufsc.br}}

\begin{abstract}
Diffusion models have recently emerged as expressive generative priors for planning and control. This paper studies Action Diffusion, an action-sequence diffusion formulation used as an open-loop proposal distribution for a point-mass system with dry friction and stiction. In this benchmark, motion starts only when the applied input exceeds a static-friction threshold, so effective controls occupy a small and temporally structured subset of the action-sequence space. A compact conditional 1D U-Net generates bounded control sequences conditioned on initial and target states. We compare it with uniform random shooting, random shooting from the same structured dataset prior, and the Cross-Entropy Method (CEM). Results show that Action Diffusion reduces terminal error and stuck steps, especially in low-sample regimes. 
These results indicate that conditional diffusion provides an effective mechanism for generating temporally coherent control sequences that overcome stiction 
by conditioning and recombining structured control primitives from the training prior for state-to-state open-loop control.
% and reliably compose motion primitives for state-to-state open-loop control.
% 
% These results indicate that conditional diffusion can provide an effective mechanism for sampling temporally coherent control sequences % \keywords{Diffusion models \and Continuous control \and Open-loop planning \and Dry friction \and Stiction}
\end{abstract}

\section{Introduction}

% Recent advances in generative modeling have opened new perspectives for sequential decision-making and control. In particular, diffusion models have been adapted to generate trajectories or action sequences conditioned on desired goals. 
Recent advances in generative modeling have opened new perspectives for sequential decision-making and control. In particular, diffusion models~\cite{Ho2020DDPM,Song2021DDIM} have been adapted to generate trajectories or action sequences conditioned on goals, rewards, or observations~\cite{Janner2022Diffuser,Ajay2023DecisionDiffuser,Chi2023DiffusionPolicy}.
While the Brazilian artificial intelligence community has recently explored diffusion models primarily for generative tasks involving static data and image processing \cite{kemmer2023performance}, their application as expressive priors for continuous physical control and action-sequencing remains largely underexplored. To bridge this gap, this paper studies an Action Diffusion formulation for control problems.

% These methods are attractive because they can represent diverse temporal patterns and can be used as proposal distributions inside sampling-based planning procedures.

Most empirical studies of diffusion-based control focus on high-dimensional settings, such as robotic manipulation or offline reinforcement-learning benchmarks~\cite{Janner2022Diffuser,Ajay2023DecisionDiffuser,Chi2023DiffusionPolicy,Levine2020OfflineRL}. These environments are important, but their complexity often makes it difficult to inspect the generated control signals and to identify why a generative prior helps. A complementary approach is to study simpler systems in which the relevant physical structure is explicit.

This paper follows that direction and considers a one-dimensional point-mass system with dry friction and stiction. Stiction creates a dead zone: when the system is nearly at rest, inputs below the static-friction threshold do not initiate motion. From a sampling perspective, this is challenging because effective control sequences must have the correct sign, magnitude, and timing. Uniformly sampled sequences may exceed the threshold occasionally, but often do so in the wrong direction or for an inappropriate duration. Dataset-prior sampling can include useful high-amplitude segments, but without conditioning it may still generate sequences suited to a different target displacement.

We study this problem through an action-sequence diffusion formulation for state-to-state open-loop control. Instead of denoising full state-action trajectories as in \cite{Janner2022Diffuser}, the model represents a conditional distribution over control sequences,
\(p_\theta(u_{0:H-1}\mid x_0,x^\star)\), where \(x_0\) is the initial state and \(x^\star\) is the desired terminal state. 
Candidate control sequences sampled from this distribution are rolled out through the known dry-friction dynamics and ranked by terminal error. This separates the generative modeling of useful control patterns from the physical evaluation of the resulting trajectories, and makes the diffusion model a learned proposal distribution for sampling-based open-loop planning. We refer to this formulation as \emph{Action Diffusion}, since diffusion is applied directly to the action/control sequence rather than to the full state-action trajectory.

% The goal is not to propose a new diffusion architecture or a complete feedback controller. Instead, the paper asks a narrower question: 
% This paper asks whether conditioning a diffusion model on the initial and target states provides a more sample-efficient proposal distribution for non-smooth open-loop control when the training prior is structured but unconditioned. 
% Since this prior already contains ``kick'', medium-amplitude, and near-zero segments, the model is not claimed to discover these primitives from scratch; the relevant question is whether conditioning places probability mass on sequences with appropriate direction, magnitude, and timing. 
% % 
% We compare Action Diffusion with uniform random shooting, dataset-prior random shooting, and CEM, where the dataset-prior baseline isolates the effect of conditioning and CEM provides a stronger iterative optimization baseline.
This paper asks whether conditioning a diffusion model on the initial and target
states provides a more sample-efficient proposal distribution for non-smooth
open-loop control when the training prior is structured but unconditioned. Our
hypothesis is that conditioning this prior on $(x_0,x^\star)$ increases the
probability of sampling sequences with suitable sign, magnitude, and timing.
Since this prior already contains ``kick'', medium-amplitude, and near-zero
segments, the model is not claimed to discover these primitives from scratch.
% 
% This paper seeks to answer the following: when a structured but unconditioned training prior is available, does conditioning a diffusion model on the initial and target states provide a more sample-efficient proposal distribution for non-smooth open-loop control? This framing also makes the role of the dataset prior explicit. Since the prior already contains ``kick'', medium-amplitude, and near-zero segments, the model is not claimed to discover these primitives from scratch. The relevant question is whether conditioning helps place probability mass on sequences with appropriate direction, magnitude, and timing.
% 
% We compare diffusion-based generation against three sampling-based baselines: uniform random shooting, dataset-prior random shooting, and the Cross-Entropy Method (CEM). The dataset-prior baseline samples from the same control prior used to generate the training data and therefore isolates the effect of conditional generation. CEM provides a stronger iterative optimization baseline and is used to study error--compute trade-offs.
% 

The main contributions of this work are:
\begin{itemize}
\item A compact action-sequence diffusion formulation for state-to-state open-loop control, in which the model learns \(p_\theta(u_{0:H-1}\mid x_0,x^\star)\) and is used as a conditional proposal distribution whose samples are evaluated by rollout through the known dry-friction dynamics.
    % \item An empirical study of conditional diffusion models as open-loop control-sequence proposal distributions for a non-smooth dry-friction system with stiction.
    \item A comparison against uniform random shooting, dataset-prior random shooting, and CEM, highlighting the difference between an unconditioned structured prior and a learned conditional prior.
    \item An analysis of the generated controls showing that the model conditions and recombines structured control segments into interpretable threshold-crossing, motion, and settling phases.
    % \item A discussion of the limitations of the current benchmark, including its synthetic nature, low dimensionality, structured training prior, and absence of feedback during execution.
\end{itemize}

\section{Related Work}

Diffusion models have recently been used as generative priors for planning and control. Diffuser~\cite{Janner2022Diffuser} denoises full state-action trajectories and uses sampling to select useful behaviors, while Decision Diffuser~\cite{Ajay2023DecisionDiffuser} formulates decision making as conditional generative modeling with conditioning variables such as returns or constraints. 
Diffusion Policy~\cite{Chi2023DiffusionPolicy} generates action sequences for visuomotor
robotic manipulation, and diffusion-based behavior cloning has also been
investigated for autonomous driving in simulated environments
\cite{Machado2025}.
Our study is complementary to these works: instead of targeting high-dimensional benchmarks, we focus on a low-dimensional non-smooth system where the temporal structure of the generated controls can be inspected directly. The main modeling distinction is that Action Diffusion denoises only the control sequence and uses the known dynamics externally for rollout and ranking, rather than generating full state-action trajectories.

The proposed evaluation is also related to sampling-based control. Random shooting generates candidate control sequences and selects the best after rollout, while adaptive methods such as CEM~\cite{Rubinstein1999CEM} iteratively refine the proposal distribution. MPPI-style methods~\cite{Williams2017MPPI} are widely used because they can handle non-convex objectives and do not require differentiability. In systems with stiction, however, large regions of the action space are ineffective because inputs below the static-friction threshold do not move the system. This makes the choice of proposal distribution central: uniform sampling, dataset-prior sampling, CEM, and conditional diffusion differ mainly in how they allocate samples in the control-sequence space.

Finally, friction is a classical source of nonlinearity and non-smooth behavior in mechanical systems~\cite{Armstrong1994FrictionSurvey,Olsson1998FrictionModels}. Static friction introduces a dead zone in which small inputs produce no motion, and classical approaches often use explicit friction models, compensation terms, or feedback design, including models such as LuGre~\cite{CanudasDeWit1995FrictionCompensation}. The present work does not replace these methods; it uses dry friction with stiction as an interpretable benchmark for studying whether a conditional generative prior can provide useful open-loop candidate sequences.

\section{Method}

\subsection{Action Diffusion Model for Control Sequences}

The proposed Action Diffusion model learns a conditional distribution over control sequences of horizon $H$,
\begin{equation}
    p_\theta\!\left(u_{0:H-1}\mid x_0, x^\star\right),
    \label{eq:cond_dist}
\end{equation}
where $x_0$ is the initial state, $x^\star$ is the desired target state, and $u_{0:H-1}$ is a candidate open-loop control sequence.
Training data are generated synthetically. Initial states are sampled from a bounded range, candidate control sequences are sampled from the training control prior described in Sect.~\ref{sec:dataset_collection}, and the known dynamics are rolled forward to obtain the terminal state $x_H$. The conditioning vector is
\begin{equation}
    c = [x_0^\top, x_H^\top]^\top \in \mathbb{R}^4 .
\end{equation}
The denoising network is then trained to model control sequences consistent with the transition from $x_0$ to $x_H$.

\subsection{Training Objective}

The model is trained with the standard noise-prediction objective used in diffusion models~\cite{Ho2020DDPM}. Given a clean control sequence $u_0$, a diffusion time index $t$, and Gaussian noise $\epsilon$, a noisy sequence is formed as
\begin{equation}
    u_t = \sqrt{\bar{\alpha}_t}\,u_0 + \sqrt{1-\bar{\alpha}_t}\,\epsilon,
    \label{eq:forward_diffusion}
\end{equation}
where $\beta_t$ is the variance increment of the forward diffusion schedule, $\alpha_t=1-\beta_t$, and $\bar{\alpha}_t=\prod_{i=1}^{t}\alpha_i$. In the experiments, the sequence of $\beta_t$ values is induced by a cosine noise schedule. The denoiser predicts the injected noise and is trained with mean-squared error:
\begin{equation}
    \mathcal{L}(\theta) =
    \mathbb{E}_{u_0,c,\epsilon,t}
    \left[\|\epsilon - \epsilon_\theta(u_t,t,c)\|_2^2\right].
    \label{eq:noise_mse}
\end{equation}

The implementation adopts classifier-free guidance~\cite{HoSalimans2022CFG} by randomly dropping the conditioning vector $c$ during training with probability $p_{\mathrm{drop}}=0.2$. During sampling, the conditional prediction uses the specified initial and target states, whereas the unconditional prediction is obtained by replacing the condition with the null condition learned during these dropped-condition training examples. The two predictions are then combined as described below.

\subsection{Model Architecture}

Each scalar control sequence $u_{0:H-1}\in\mathbb{R}^{H}$ is represented as a tensor with one channel and $H$ temporal positions, i.e., shape $1\times H$ with $H=64$ in the experiments. The denoiser is a compact 1D conditional U-Net that maps this noisy one-channel temporal signal to a noise prediction with the same shape. A sinusoidal embedding encodes the diffusion time step, while the conditioning vector $[x_0,x_H]$ is mapped through a small MLP and injected additively into each residual block. Table~\ref{tab:unet} summarizes the implemented architecture used in the experiments.
%which matches the Python implementation used in the experiments.

\begin{table}[t]
\caption{Conditional 1D U-Net used to denoise control sequences of length $H=64$. All residual blocks use GroupNorm, SiLU activations, and additive conditioning from time and state embeddings.}
\label{tab:unet}
\centering
\footnotesize
\begin{tabular}{@{}lll@{}}
\toprule
Stage & Layer(s) & Output shape \\ \midrule
Input & Control sequence reshaped to 1 channel & $1 \times 64$ \\
Embeddings & Sinusoidal time embedding $\rightarrow$ MLP; condition MLP & $64$, $64$ \\
Encoder & Conv1D $(1 \rightarrow 64)$, kernel $3$ & $64 \times 64$ \\
 & ResBlock1D & $64 \times 64$ \\
 & ResBlock1D & $64 \times 64$ \\
Downsampling & Conv1D, kernel $4$, stride $2$ & $64 \times 32$ \\
Bottleneck & ResBlock1D & $64 \times 32$ \\
 & ResBlock1D & $64 \times 32$ \\
Upsampling & Transposed Conv1D, kernel $4$, stride $2$ & $64 \times 64$ \\
Decoder & Skip addition from encoder, ResBlock1D & $64 \times 64$ \\
 & ResBlock1D & $64 \times 64$ \\
Output & Conv1D $(64 \rightarrow 1)$, kernel $3$ & $1 \times 64$ \\ \bottomrule
\end{tabular}
\end{table}

\subsection{Sampling with DDIM and Classifier-Free Guidance}

At test time, control sequences are sampled with a deterministic DDIM-style reverse process~\cite{Song2021DDIM}. Starting from Gaussian noise, the sampler predicts the denoised control sequence over a reduced subset of diffusion steps. For each reverse step, the implementation computes unconditional and conditional noise predictions and combines them as
\begin{equation}
    \hat{\epsilon}_{\mathrm{cfg}} =
    \hat{\epsilon}_{\mathrm{uncond}} +
    s\left(\hat{\epsilon}_{\mathrm{cond}} - \hat{\epsilon}_{\mathrm{uncond}}\right),
\end{equation}
where $s$ is the guidance scale. Here, $\hat{\epsilon}_{\mathrm{cond}}$ denotes the prediction with the actual condition $c=[x_0^\top,x^{\star\top}]^\top$, while $\hat{\epsilon}_{\mathrm{uncond}}$ denotes the prediction under the null condition. The predicted clean sequence is clamped during sampling for numerical stability.

\subsection{Open-Loop Control}
\label{sec:open_loop_control}

For a target state $x^\star$, each method generates $K$ candidate open-loop sequences. The candidates are rolled out through the known dynamics, and the best sequence is selected by terminal state error,
\begin{equation}
    e_{\mathrm{final}} = \|x_H - x^\star\|_2 .
    \label{eq:terminal_error_select}
\end{equation}
The same selection rule is used for Action Diffusion, dataset-prior random shooting, uniform random shooting, and for ranking samples within each CEM iteration.

For a multi-step reference, we use segmented open-loop planning. Given a sequence of target states $(x_1^\star,\ldots,x_M^\star)$, the method generates a sequence from the current state to $x_1^\star$, executes the selected segment, then repeats the procedure from the reached state to $x_2^\star$, and so on. This is not an online feedback controller: it is a piecewise composition of state-to-state open-loop plans.

\section{Experimental Setup}

\subsection{Benchmark: Dry Friction with Stiction}

We evaluate the approach on a one-dimensional point-mass system with state $x=[p,v]^\top$ and scalar control $u$. The dynamics combine Coulomb friction, viscous damping, and a static-friction dead zone:
\begin{equation}
\dot{p} = v,
\end{equation}
\begin{equation}
\dot{v} =
\begin{cases}
0, & |v| < \epsilon_v \text{ and } |u| \le f_s, \\
u - f_c\operatorname{sgn}(u), & |v| < \epsilon_v \text{ and } |u| > f_s, \\
u - f_c\operatorname{sgn}(v) - b v, & |v| \ge \epsilon_v .
\end{cases}
\label{eq:stiction_dynamics}
\end{equation}
The first case models stiction: when the velocity is close to zero and the input does not exceed the static-friction threshold, the mass remains at rest. The system is simulated with a semi-implicit Euler update, using the post-friction velocity to update the position.

\subsection{Dataset Collection}
\label{sec:dataset_collection}

The dataset consists of synthetic pairs $(u_{0:H-1},c)$, where $c=[x_0^\top,x_H^\top]^\top$. Initial states are sampled uniformly from the configured ranges. No measurement noise or process noise is injected in the current benchmark; every terminal state is obtained by rolling out the known dry-friction dynamics.

For the dry-friction benchmark, the control sequence distribution is intentionally not uniform. Instead, we use a structured prior designed to contain threshold-crossing segments that can initiate motion under stiction. Each sequence is composed of eight constant segments. For each segment, one of three modes is sampled: a high-amplitude ``kick'' mode with magnitude uniformly sampled in $[0.75u_{\max},u_{\max}]$ and random sign; a medium-amplitude mode sampled from $[-0.6u_{\max},0.6u_{\max}]$; and a near-zero mode sampled from $[-0.15u_{\max},0.15u_{\max}]$. The segment values are expanded to length $H$ and smoothed by a first-order filter with coefficient $\alpha=0.3$. The final sequence is clipped to $[-u_{\max},u_{\max}]$.

% This design is important for interpreting the results. Action Diffusion is not trained from an unstructured action distribution; the training prior already contains high-amplitude, medium-amplitude, and near-zero primitives. The role of the learned model is therefore to condition this structured prior on the transition from $x_0$ to $x_H$, assigning probability to sequences with suitable sign, amplitude, and timing. The dataset-prior random shooting baseline uses the same control prior but does not condition samples on the target state.

The role of the learned model is therefore to condition this structured prior on the transition from $x_0$ to $x_H$, assigning probability to sequences with suitable sign, amplitude, and timing. The dataset-prior random shooting baseline uses the same control prior but does not condition samples on the target state.

\subsection{Parameter Configuration}

Table~\ref{tab:param_cfg} summarizes the main configuration used in the experiments. The Action Diffusion model was trained on $50{,}000$ training transitions and $10{,}000$ validation transitions. Training was run for $15{,}000$ gradient steps using Adam and an exponential moving average of the model weights.

\begin{table}[t]
\caption{Main parameter configuration used for training and evaluation.}
\label{tab:param_cfg}
\centering
\footnotesize
\begin{tabular}{@{}ll@{}}
\toprule
Component & Value \\ \midrule
% System & Dry friction with stiction \\
Horizon $H$ & 64 \\
Sampling time $\Delta t$ & 0.05 \\
Control bound $u_{\max}$ & 0.9 \\
Static friction $f_s$ & 0.7 \\
Coulomb friction $f_c$ & 0.6 \\
Viscous damping $b$ & 0.05 \\
Velocity threshold $\epsilon_v$ & $10^{-3}$ \\
Initial-state range & $p_0\in[-1,1]$, $v_0\in[-0.5,0.5]$ \\
Training / validation samples & $50{,}000$ / $10{,}000$ \\
Diffusion steps $T$ & 200 \\
DDIM steps & 50 \\
Noise schedule & Cosine \\
Batch size & 256 \\
Training steps & $15{,}000$ \\
Learning rate & $2\times10^{-3}$ \\
Condition drop probability $p_{\mathrm{drop}}$ & 0.2 \\
Base channels / time dim / condition dim & 64 / 64 / 64 \\
EMA decay & 0.995 \\
Guidance scale & 2.0 \\
Multi-step reference sequence & $(0.4,0)$, $(-0.4,0)$, $(0.2,0)$ \\
\bottomrule
\end{tabular}
\end{table}

\subsection{CEM Baseline}

In addition to random shooting baselines, we evaluate the Cross-Entropy Method (CEM) as a stronger sampling-based optimizer. For a given target state, CEM maintains a Gaussian distribution over control sequences and iteratively refines its mean and variance using the elite candidates with lowest terminal error. At each iteration, candidate sequences are sampled, clipped to the admissible control range $[-u_{\max},u_{\max}]$, rolled out through the dry-friction dynamics, and ranked according to the terminal error in Eq.~\eqref{eq:terminal_error_select}. The updated sampling distribution is then fitted to the elite set.

To assess the trade-off between optimization quality and computational cost, we vary the number of CEM iterations while keeping the candidate population per iteration fixed within each operating point. We then compare the resulting terminal error against the wall-clock compute time required by CEM and by Action Diffusion sampling. The comparison is performed for two candidate budgets, $K=32$ and $K=512$, under the same reference and initial condition.

\subsection{Evaluation Metrics}

For a generated sequence, we report terminal error $e_{\mathrm{final}}$ as defined in Eq.~\eqref{eq:terminal_error_select}. We also report control energy,
\begin{equation}
    E_u = \sum_{k=0}^{N-1} u_k^2,
\end{equation}
and control smoothness,
\begin{equation}
    S_u = \sum_{k=0}^{N-2} (u_{k+1}-u_k)^2.
\end{equation}
For the dry-friction system, we additionally report the number of stuck steps,
\begin{equation}
    N_{\mathrm{stuck}} = \sum_k \mathbf{1}\{ |v_k|<\epsilon_v,\ |u_k|<f_s,\ |p_k-p_k^\star|>\delta\},
\end{equation}
with $\delta=0.01$. This metric counts steps in which the system is effectively stopped, the applied input is below the static-friction threshold, and the position is still away from the desired reference. For multi-step experiments, stuck steps and computation time are computed over the composed open-loop trajectory. We report both the final error, measured only with respect to the last target after all segments have been executed, and the mean end error, defined as the average terminal error over all segment endpoints.

\section{Results}

% TODO EXPERIMENT: Add CEM to the main K-sweep for K in {16, 32, 128, 512}, using the same number of independent runs as the other methods.
% TODO EXPERIMENT: Add a random target-distribution evaluation over many (x0, x*) pairs sampled from the test support, with success rate and median/mean terminal error.
% TODO EXPERIMENT: Add a compact sensitivity analysis over fs, fc, b, and/or umax.

\subsection{Single-Target Sample Budget}

We begin with a small target displacement, $x^\star=(0.2,0)$, which is particularly challenging due to the presence of static friction. In this regime, the controller must generate sufficient force to overcome stiction while avoiding excessive overshoot. As a result, successful behavior depends on producing temporally coordinated control sequences, rather than simply low-energy inputs.

\begin{figure}
    \centering
    \includegraphics[width=0.7\linewidth]{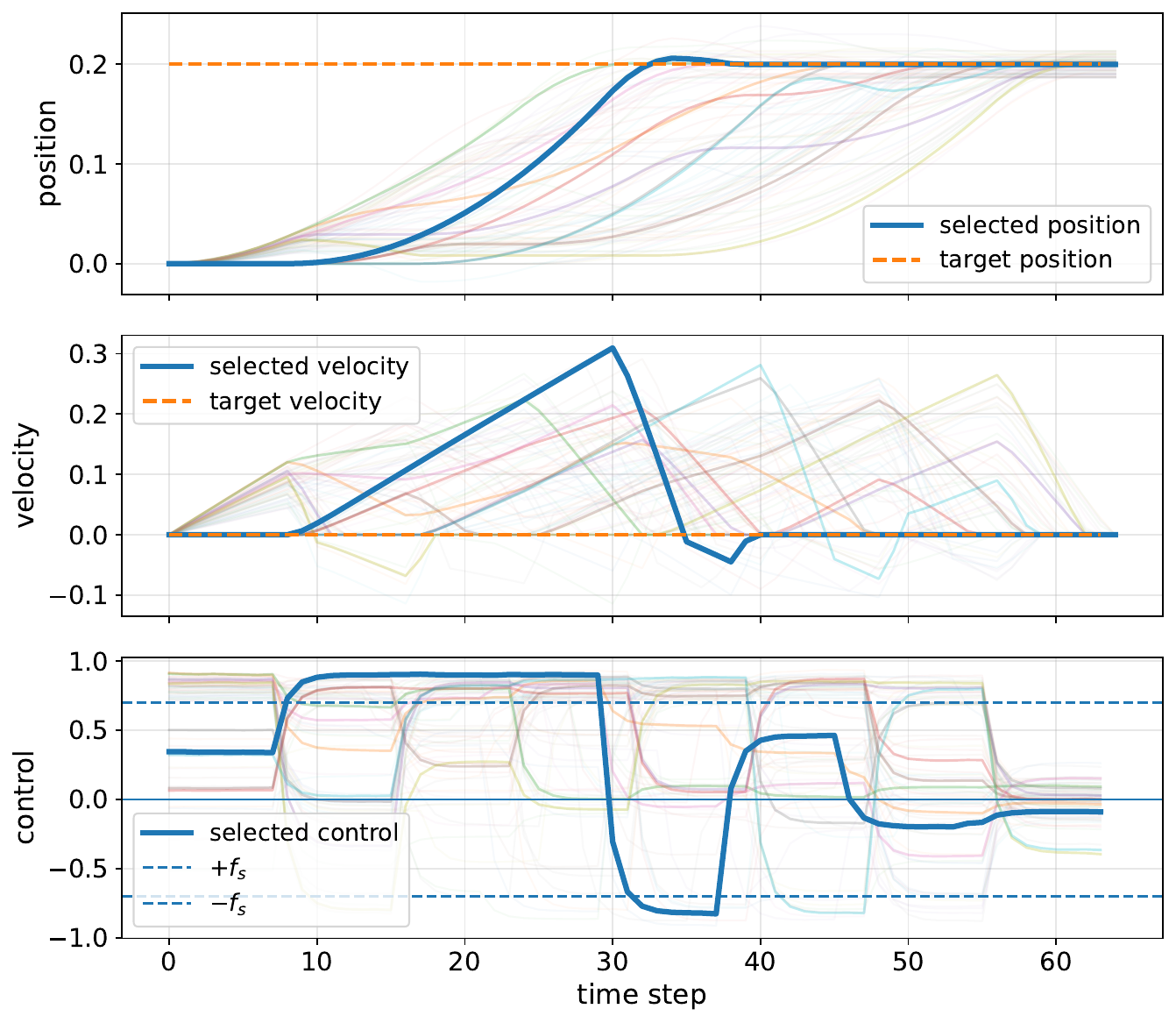}
    \caption{Action Diffusion open-loop control for the small target $x^\star=(0.2,0)$ using $K=512$ sampled trajectories. Faint lines represent candidate trajectories, and the highlighted line corresponds to the selected best sequence.}
    \label{fig:target02}
\end{figure}

Figure~\ref{fig:target02} provides a qualitative illustration of Action Diffusion open-loop control with $K=512$ sampled trajectories. Each faint trajectory corresponds to a candidate control sequence, while the highlighted trajectory represents the selected best sample. The figure shows that the Action Diffusion proposal concentrates many candidates near trajectories that approach the target. In particular, the selected trajectory exhibits a clear threshold-crossing and settling pattern: the control signal initially exceeds the static-friction threshold to initiate motion and is subsequently modulated to regulate the system near the target. Since the training prior already contains high-amplitude and near-zero segments, this behavior should be interpreted as conditional selection and timing of available primitives, rather than as discovery of the primitive structure from scratch.

To quantify this behavior, for each method and sample budget $K\in\{16,32,128,512\}$, we perform 10 independent runs. In each run, $K$ candidate control sequences are sampled, and the best sequence is selected according to the terminal error.

\begin{figure}[t]
\centering
\includegraphics[width=1\linewidth]{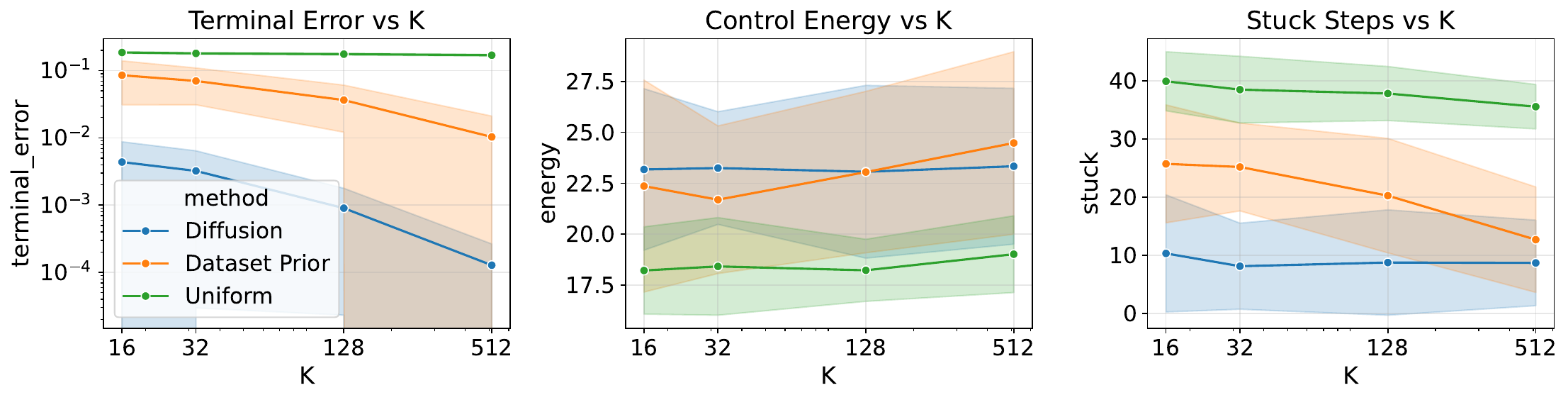}
\caption{Performance as a function of the sample budget $K\in\{16,32,128,512\}$. Results show mean and standard deviation over 30 independent runs.}
\label{fig:k_sweep}
\end{figure}

Figure~\ref{fig:k_sweep} reports the resulting terminal error, control energy, and number of stuck steps. Action Diffusion achieves low terminal error even for small $K$, indicating that the learned conditional distribution concentrates probability mass on controls that are appropriate for the requested transition. Dataset-prior random shooting samples from the same structured prior, but because it is not conditioned on the target, many samples contain kicks with unsuitable sign, duration, or timing. Uniform random shooting remains largely ineffective across all values of $K$ because most randomly sampled sequences either fail to cross the static-friction threshold at the right time or produce poorly coordinated motion.

The energy results further highlight that low energy alone is not sufficient for successful control. In particular, a method can have low control energy simply because it does not move the system effectively. For this reason, terminal error and stuck steps are more informative than energy alone in this benchmark.

% # TODO: explain that in the plot above, we did not include CEM because for a given K, there is n_iters optimization iterations. Thus, the budget means something different for CEM.

\subsection{Error--Compute Trade-off Against CEM}

\begin{figure}[t]
\centering
% \subfigure[K=16]{
% \includegraphics[width=0.47\linewidth]{figures/pareto_cem_diffusion_std_k16.pdf} 
% }
\subfigure[K=32]{
\includegraphics[width=0.47\linewidth]{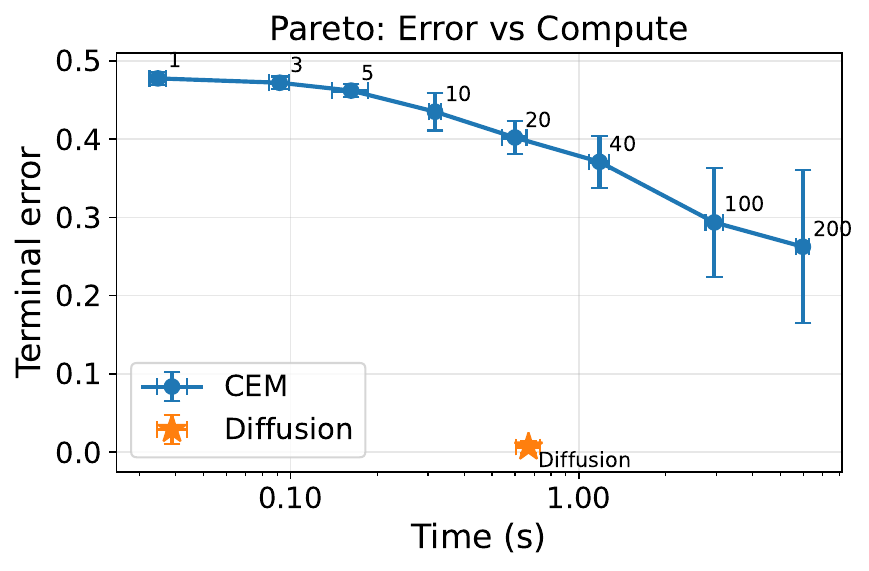} 
}
% \subfigure[K=128]{
% \includegraphics[width=0.47\linewidth]{figures/pareto_cem_diffusion_std_k128.pdf}
% }
\subfigure[K=512]{
\includegraphics[width=0.47\linewidth]
{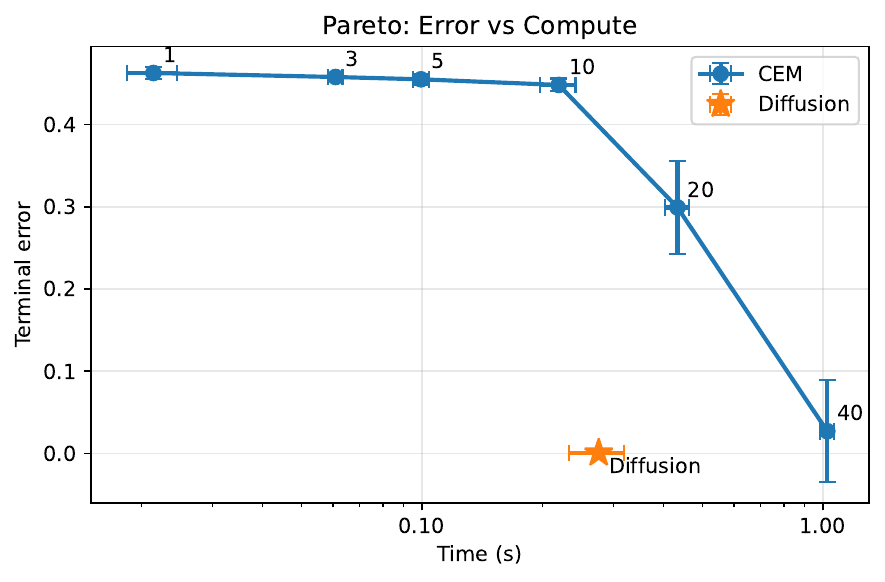}
}
\caption{
Error--compute comparison between Action Diffusion and CEM for representative low- and high-budget regimes. CEM points correspond to different numbers of optimization iterations. Action Diffusion is shown as a single operating point. Lower and leftward points are preferable.
}
% \caption{Pareto comparison between terminal error and compute time for Action Diffusion and CEM considering the fixed reference $(0.5,0)$ and initial condition $(0,0)$. 
% % The four panels show the comparison for $K\in\{16,32,128,512\}$. 
% For CEM, each point corresponds to a different number of optimization iterations. 
% % Action Diffusion is shown as a single operating point in each panel. Lower and leftward points are preferable. 
% Thirty runs are computed per configuration, with the resulting mean and standard deviation shown in the plot.}
\label{fig:pareto_cem_diffusion}
\end{figure}

We also compare Action Diffusion with CEM under a fixed target-setting of $(0.5,0)$ and initial condition $(0,0)$, focusing on the trade-off between terminal accuracy and computation time. 
Figure~\ref{fig:pareto_cem_diffusion} reports representative low- and high-budget cases, $K=32$ and $K=512$, where
thirty runs are computed per configuration, with the resulting mean and standard deviation shown in the plot.
For CEM, each point corresponds to a different number of optimization iterations, while Action Diffusion is shown as a single operating point.
The comparison shows that the advantage of Action Diffusion depends on the operating regime. For $K=32$, Action Diffusion achieves lower terminal error at a smaller compute budget than CEM. For $K=512$, the gap becomes narrower: CEM improves with additional refinement, and the advantage of Action Diffusion is less pronounced.

% We also compare Action Diffusion with the Cross-Entropy Method (CEM) under a fixed target-setting, focusing on the trade-off between terminal accuracy and computation time.
% % 
% Figure~\ref{fig:pareto_cem_diffusion} reports terminal error versus wall-clock time for both methods for $K\in\{16,32,128,512\}$. For CEM, each point corresponds to a different number of optimization iterations, yielding progressively better solutions at increasing computational cost. Action Diffusion is shown as a single operating point in each panel.

% The comparison shows that the advantage of Action Diffusion depends on the operating regime. For $K=32$, Action Diffusion achieves substantially lower terminal error at a smaller compute budget than CEM. For $K=512$, the gap becomes narrower: CEM improves significantly with additional refinement, and although Action Diffusion still attains essentially zero error, it is no longer uniformly better in both objectives by the same margin observed in the low-budget regime.

% \subsection{Segmented Multi-Step Open-Loop Planning}

% \subsubsection{Candidate Structure}

% Figure~\ref{fig:candidates} illustrates the set of candidate trajectories generated by Action Diffusion for a single run. Each segment is produced by sampling $K$ control sequences conditioned on the current state and target. The transparent trajectories show the diversity of candidate rollouts, while the bold trajectory corresponds to the selected sequence.

% \begin{figure}[t]
% \centering
% \includegraphics[width=0.95\linewidth]{figures/open_loop_step_reference_candidates_20260430_203246.pdf}
% \caption{
% Candidate trajectories generated by Action Diffusion for the segmented multi-step task under dry friction. Transparent lines correspond to sampled candidates, while the bold trajectory represents the selected sequence based on terminal error.
% }
% \label{fig:candidates}
% \end{figure}

\subsection{Segmented Multi-Step Open-Loop Planning}

We next evaluate the segmented open-loop task with the step-reference sequence $(0.4,0)$, $(-0.4,0)$, and $(0.2,0)$. In each segment, Action Diffusion samples candidate control sequences conditioned on the current state and the next target, rolls them out through the known dynamics, and selects the best one by terminal error.

% Figure~\ref{fig:multistep_runs} compares Action Diffusion, dataset-prior random shooting, and CEM for $K=16$ and $K=512$. For each method and value of $K$, five independent selected trajectories are plotted. With $K=16$, Action Diffusion follows the intended sequence more consistently, whereas dataset-prior random shooting shows higher variability. With $K=512$, the dataset-prior baseline improves but remains less consistent. The control profiles show that Action Diffusion tends to place threshold-crossing segments near the portions of the trajectory where motion must be initiated, while the unconditioned prior does not know which direction or timing is appropriate for the current target.

\begin{figure}[t]
\centering
\includegraphics[width=1\linewidth]{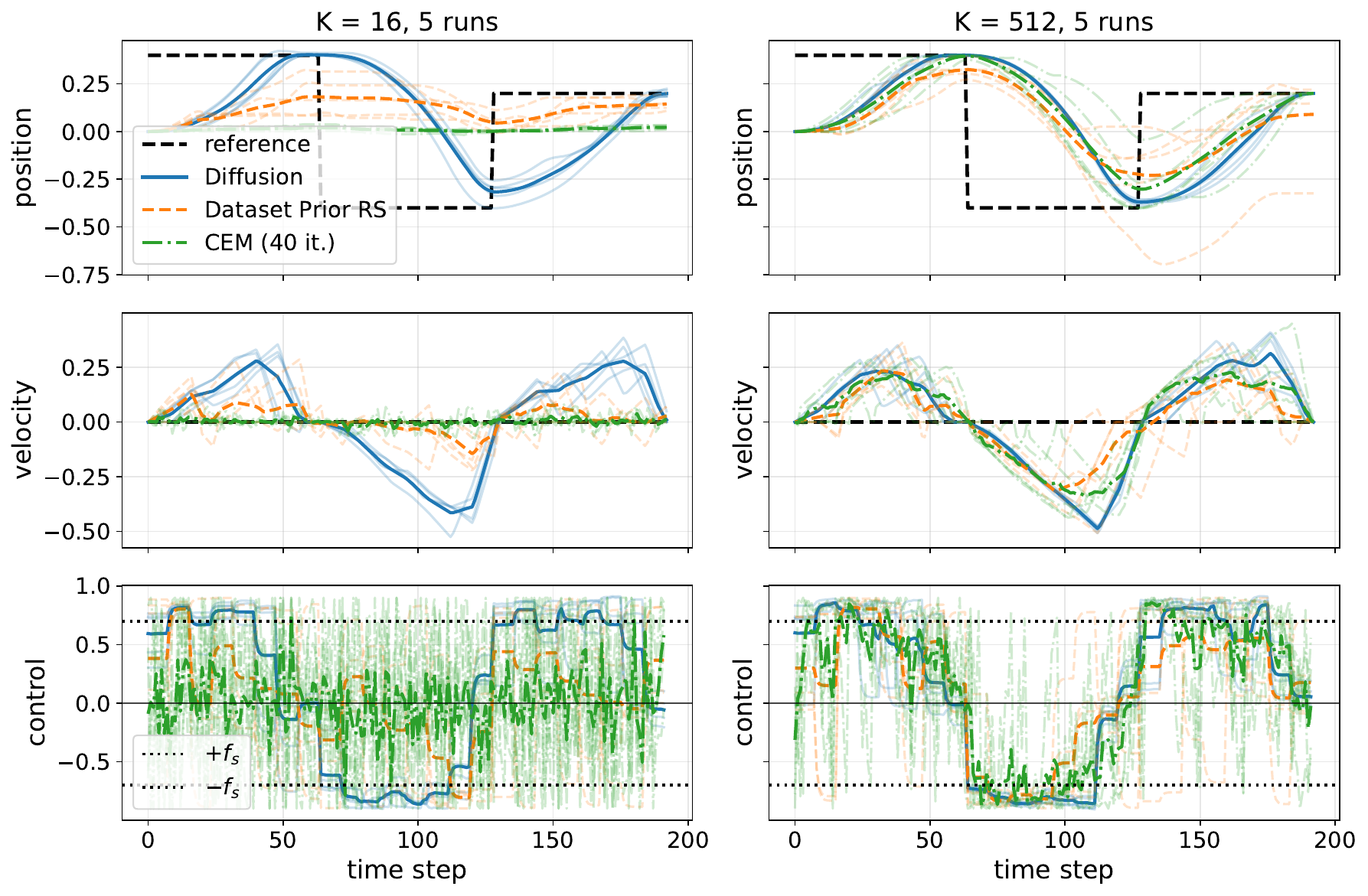}
\caption{Segmented open-loop trajectories for the multi-step control task under dry friction. 
Results are shown for $K=16$ and $K=512$. For each method and value of $K$, 5 independent runs are plotted with thin transparent lines, while the mean of them is given by a solid thick line.
}
\label{fig:multistep_runs}
\end{figure}

Figure~\ref{fig:multistep_runs} compares Action Diffusion, dataset-prior random shooting, and CEM for $K=16$ and $K=512$. 
For each method and value of $K$, five independent selected trajectories are plotted, with the thick line showing their mean.
With $K=16$, Action Diffusion follows the intended sequence more consistently, whereas CEM remains almost flat and fails to initiate the required motion. This illustrates the difficulty of iterative refinement when the initial sampled population contains few effective threshold-crossing controls.
With $K=512$, CEM improves on average, but several individual runs still deviate substantially from the reference, indicating high variability in the non-smooth stiction landscape. Action Diffusion produces more consistent threshold-crossing behavior across runs because its proposal distribution is conditioned on the desired state transition.

% Figure~\ref{fig:multistep_runs} compares Action Diffusion, dataset-prior random shooting, and CEM for $K=16$ and $K=512$. 
% For each configuration of $K$, five independent runs are plotted. For each run, the best trajectory is selected from $K$ sampled candidates based on terminal error.

% With $K=16$, Action Diffusion consistently produces trajectories that follow the intended sequence, whereas dataset-prior random shooting shows much higher variability. With $K=512$, the dataset-prior baseline improves, but its trajectories remain less consistent. The control profiles further show that Action Diffusion tends to place threshold-crossing segments near the portions of the trajectory where motion must be initiated. This is precisely where conditioning is expected to help: the unconditioned dataset prior contains similar segment types, but it does not know which direction or timing is appropriate for the current target.

% \begin{figure}[t]
% \centering
% \includegraphics[width=1\linewidth]{figures/step_reference_runs_K16_K512_x2.pdf}
% \caption{Segmented open-loop trajectories for the multi-step control task under dry friction. Results are shown for $K=16$ and $K=512$. For each method and value of $K$, 5 independent runs are plotted, where the best trajectory is selected from $K$ sampled candidates based on terminal error.}
% \label{fig:multistep_runs}
% \end{figure}
% 

% \subsubsection{Quantitative Results}

\paragraph{Quantitative results.}
Table~\ref{tab:multistep_results} summarizes the randomized multi-step evaluation over 30 reference sequences and 5 independent sampling runs per sequence. Action Diffusion obtains the lowest final error, mean end error, and number of stuck steps for all values of $K$. Dataset-prior sampling is faster because it only samples and rolls out candidate controls, but its larger errors show that the unconditioned structured prior is not sufficient. CEM improves as the population size increases, but it requires substantially more computation under the 40-iteration setting.
\begin{table}[h!]
\caption{
Comparison of methods 
on multi-step open-loop control. 
% under dry friction with stiction. 
% Results are reported as mean $\pm$ standard deviation over 10 runs. For each run, $K$ candidate sequences are sampled and the best is selected based on terminal error.
Results are reported as mean $\pm$ standard deviation over 30 randomly sampled multi-step reference sequences and 5 independent sampling runs per sequence, totaling 150 evaluations per method and sample budget $K$.
}
\label{tab:multistep_results}
\centering
\footnotesize
\begin{tabular}{lccccc}
\toprule
Method & K & Final Error $\downarrow$ & Mean End Error $\downarrow$ & Stuck Steps $\downarrow$ & Time (s) $\downarrow$ \\
\midrule
Action Diffusion & 16 & \textbf{0.20 $\pm$ 0.3} & \textbf{0.17 $\pm$ 0.1} & \textbf{23.2 $\pm$ 18.7} & 0.68 $\pm$ 0.1 \\
Dataset Prior & 16 & 0.38 $\pm$ 0.4 & 0.40 $\pm$ 0.2 & 69.9 $\pm$ 19.3 & \textbf{0.07 $\pm$ 0.0} \\
CEM (40 it.) & 16 & 0.43 $\pm$ 0.3 & 0.49 $\pm$ 0.2 & 105.6 $\pm$ 11.3 & 2.54 $\pm$ 0.2 \\
\midrule
Action Diffusion & 32 & \textbf{0.19 $\pm$ 0.3} & \textbf{0.16 $\pm$ 0.1} & \textbf{18.8 $\pm$ 18.1} & 0.75 $\pm$ 0.1 \\
Dataset Prior & 32 & 0.34 $\pm$ 0.4 & 0.37 $\pm$ 0.2 & 60.7 $\pm$ 17.6 & \textbf{0.08 $\pm$ 0.0} \\
CEM (40 it.) & 32 & 0.36 $\pm$ 0.3 & 0.40 $\pm$ 0.2 & 63.0 $\pm$ 9.3 & 2.57 $\pm$ 0.3 \\
\midrule
Action Diffusion & 128 & \textbf{0.18 $\pm$ 0.3} & \textbf{0.15 $\pm$ 0.1} & \textbf{17.1 $\pm$ 14.9} & 0.76 $\pm$ 0.1 \\
Dataset Prior & 128 & 0.32 $\pm$ 0.4 & 0.31 $\pm$ 0.2 & 46.5 $\pm$ 16.6 & \textbf{0.11 $\pm$ 0.0} \\
CEM (40 it.) & 128 & 0.31 $\pm$ 0.4 & 0.25 $\pm$ 0.2 & 27.9 $\pm$ 16.9 & 2.56 $\pm$ 0.1 \\
\midrule
Action Diffusion & 512 & \textbf{0.17 $\pm$ 0.3} & \textbf{0.13 $\pm$ 0.1} & \textbf{15.9 $\pm$ 17.0} & 1.56 $\pm$ 0.6 \\
Dataset Prior & 512 & 0.29 $\pm$ 0.4 & 0.25 $\pm$ 0.2 & 33.6 $\pm$ 15.7 & \textbf{0.33 $\pm$ 0.2} \\
CEM (40 it.) & 512 & 0.19 $\pm$ 0.3 & 0.14 $\pm$ 0.1 & 22.2 $\pm$ 16.7 & 3.47 $\pm$ 1.2 \\
\bottomrule
\end{tabular}
\end{table}
%%%%%%%%%%%%%%%%%%%%%%%%%%%%%%%%
% 
% \paragraph{Sensitivity to the static-friction threshold.}
To test whether the conclusions are specific to the nominal dry-friction configuration, we also repeated the randomized multi-step evaluation for different values of the static-friction threshold ($f_s$). Table~\ref{tab:sensitivity_fs} reports the mean end error for 
($f_s \in \{0.6,0.7,0.8\}$) 
using ($K=128$). Action Diffusion is trained from scratch for each value of $f_s$. It
achieves the lowest mean end error in all three settings, suggesting that the benefit of conditioning the action-sequence prior is not restricted to a single stiction threshold. The experiment is not intended as a complete robustness analysis, since the remaining system parameters are kept fixed, but it provides an initial sensitivity check over the main parameter governing the stiction dead zone.

%%%%%%%%%%%%%%%%%%%%%%%%%%%%%%%% 
\begin{table}[tb]
\centering
\footnotesize
\caption{Sensitivity to the static-friction threshold $f_s$. Results report the Mean End Error 
% as mean $\pm$ standard deviation over randomized multi-step reference sequences (150 evaluations per method 
(as in Table~\ref{tab:multistep_results}), for sample budget $K=128$. 
% The best result for each value of $f_s$ is shown in bold.
}
\label{tab:sensitivity_fs}
\begin{tabular}{lccc}
\toprule
Method & $f_s = 0.6$ & $f_s = 0.7$ & $f_s = 0.8$ \\
\midrule
Action Diffusion
& \textbf{0.14 $\pm$ 0.1} 
& \textbf{0.15 $\pm$ 0.1} 
& \textbf{0.10 $\pm$ 0.1} \\
Dataset Prior 
& 0.30 $\pm$ 0.2 
& 0.31 $\pm$ 0.2 
& 0.31 $\pm$ 0.2 \\
CEM (40 it.) 
& 0.22 $\pm$ 0.2 
& 0.25 $\pm$ 0.2 
& 0.22 $\pm$ 0.2 \\
\bottomrule
\end{tabular}
\end{table}

\section{Discussion}
% TODO: shorten here

The results suggest that Action Diffusion is useful as a structured open-loop proposal distribution for non-smooth control. The comparison with dataset-prior random shooting is central: both methods use the same underlying control-sequence prior, but only Action Diffusion conditions it on the desired transition. Thus, the gains should be attributed not to the mere presence of high-amplitude ``kick'' segments in the data, but to the model's ability to assign higher probability to sequences with suitable direction, magnitude, and timing. The CEM comparison further indicates that this advantage is strongest in low-budget regimes, where iterative refinement from a generic sampling distribution is inefficient due to flat or ineffective regions induced by stiction. At larger candidate budgets, CEM improves and the advantage of Action Diffusion becomes less pronounced.

% The results suggest that Action Diffusion can serve as a structured open-loop proposal distribution for non-smooth control problems. The comparison with dataset-prior random shooting is central to this interpretation. Both methods have access to the same underlying control-sequence prior, including high-amplitude, medium-amplitude, and near-zero segments. The difference is that Action Diffusion conditions this prior on the desired transition. Thus, the improvement should not be attributed simply to the existence of ``kick'' segments in the data, but to the model's ability to assign higher probability to sequences with suitable direction, magnitude, and timing.
% % 
% The comparison against CEM provides a complementary perspective. In the low-budget regime, iterative refinement from a generic sampling distribution is relatively inefficient because many candidate sequences lie in flat or ineffective regions induced by stiction. Action Diffusion, by contrast, directly samples from a learned conditional distribution adapted to the structure of successful transitions. At larger candidate budgets, CEM improves and the advantage of Action Diffusion becomes less pronounced. 

\section{Conclusion}

This paper studied Action Diffusion as an open-loop control-sequence proposal distribution for a point-mass system with dry friction and stiction. The benchmark requires inputs that cross a static-friction threshold to initiate motion, making naive sampling inefficient. The results show that Action Diffusion can achieve lower terminal error and fewer stuck steps than uniform and dataset-prior random shooting, especially with small sample budgets. The CEM comparison indicates that this advantage is strongest in low-budget regimes, while iterative optimization can reduce the gap when more computation is available. Qualitative results show interpretable threshold-crossing, motion, and settling phases, but these should be understood as conditional recombinations of structured segments already present in the training prior. Future work will consider richer mechanical systems, stochastic disturbances, additional learning baselines, and receding-horizon feedback execution.

% ---

\paragraph{Acknowledgments.}
% This study was funded by CNPq-Brazil (grant nr. XXX) and YYY (grant nr. XXX).
This work was partially funded by the National Council for Scientific and Technological Development – CNPq, Brazil (Grant No. 420148/2025-6), and FAPESC -- Foundation for Research and Innovation Support of the State of Santa Catarina (Grant No. 2024TR000090).

\paragraph{Disclosure of Interests.}
The author has no competing interests to declare that are relevant to the content of this article.

\balance
\bibliographystyle{sbc}
\bibliography{references_updated}

@inproceedings{Machado2025,
  title={Investigating diffusion models for offline behavior cloning in autonomous driving scenarios},
  author={Machado, Bruno M and Antonelo, Eric A},
  booktitle={2025 Brazilian Conference on Robotics (CROS)},
  volume={1},
  pages={1--6},
  year={2025},
  organization={IEEE}
}

@inproceedings{kemmer2023performance,
  author    = {Kemmer, B. and Sim{\~o}es, R. and Ivamoto, V. and Lima, C.},
  title     = {Performance Analysis of Generative Adversarial Networks and Diffusion Models},
  booktitle = {Proceedings of the Brazilian Conference on Intelligent Systems (BRACIS)},
  publisher = {Springer},
  year      = {2023}
}

@article{Levine2020OfflineRL,
  author = {Levine, Sergey and Kumar, Aviral and Tucker, George and Fu, Justin},
  title = {Offline Reinforcement Learning: Tutorial, Review, and Perspectives},
  journal = {arXiv preprint arXiv:2005.01643},
  year = {2020}
}

@inproceedings{Williams2017MPPI,
  author = {Williams, Grady and Drews, Paul and Goldfain, Brian and Rehg, James and Theodorou, Evangelos},
  title = {Information-Theoretic Model Predictive Control: Theory and Applications to Autonomous Driving},
  booktitle = {ICRA},
  year = {2017}
}

@article{CanudasDeWit1995FrictionCompensation,
  author={Canudas de Wit, Carlos and Olsson, Håkan and Åström, Karl Johan and Lischinsky, Pablo},
  title={A new model for control of systems with friction},
  journal={IEEE Transactions on Automatic Control},
  year={1995},
  volume={40},
  number={3},
  pages={419--425}
}

@inproceedings{Ho2020DDPM,
  title={Denoising Diffusion Probabilistic Models},
  author={Ho, Jonathan and Jain, Ajay and Abbeel, Pieter},
  booktitle={Advances in Neural Information Processing Systems},
  volume={33},
  pages={6840--6851},
  year={2020}
}

@inproceedings{Song2021DDIM,
  title={Denoising Diffusion Implicit Models},
  author={Song, Jiaming and Meng, Chenlin and Ermon, Stefano},
  booktitle={International Conference on Learning Representations},
  year={2021},
  url={https://openreview.net/forum?id=St1giarCHLP}
}

@article{HoSalimans2022CFG,
  title={Classifier-Free Diffusion Guidance},
  author={Ho, Jonathan and Salimans, Tim},
  journal={arXiv preprint arXiv:2207.12598},
  year={2022}
}

@inproceedings{Janner2022Diffuser,
  title={Planning with Diffusion for Flexible Behavior Synthesis},
  author={Janner, Michael and Du, Yilun and Tenenbaum, Joshua B. and Levine, Sergey},
  booktitle={Proceedings of the 39th International Conference on Machine Learning},
  series={Proceedings of Machine Learning Research},
  volume={162},
  pages={9902--9915},
  year={2022}
}

@inproceedings{Ajay2023DecisionDiffuser,
  title={Is Conditional Generative Modeling All You Need for Decision-Making?},
  author={Ajay, Anurag and Du, Yilun and Gupta, Abhi and Tenenbaum, Joshua B. and Jaakkola, Tommi S. and Agrawal, Pulkit},
  booktitle={International Conference on Learning Representations},
  year={2023},
  url={https://openreview.net/forum?id=sP1fo2K9DFG}
}

@inproceedings{Chi2023DiffusionPolicy,
  title={Diffusion Policy: Visuomotor Policy Learning via Action Diffusion},
  author={Chi, Cheng and Xu, Zhenjia and Feng, Siyuan and Cousineau, Eric and Du, Yilun and Burchfiel, Benjamin and Tedrake, Russ and Song, Shuran},
  booktitle={Robotics: Science and Systems},
  year={2023}
}

@article{Rubinstein1999CEM,
  title={The Cross-Entropy Method for Combinatorial and Continuous Optimization},
  author={Rubinstein, Reuven Y.},
  journal={Methodology and Computing in Applied Probability},
  volume={1},
  number={2},
  pages={127--190},
  year={1999},
  doi={10.1023/A:1010091220143}
}

@article{Armstrong1994FrictionSurvey,
  title={A Survey of Models, Analysis Tools and Compensation Methods for the Control of Machines with Friction},
  author={Armstrong-H{\'e}louvry, Brian and Dupont, Pierre and Canudas de Wit, Carlos},
  journal={Automatica},
  volume={30},
  number={7},
  pages={1083--1138},
  year={1994},
  doi={10.1016/0005-1098(94)90209-7}
}

@article{Olsson1998FrictionModels,
  title={Friction Models and Friction Compensation},
  author={Olsson, Henrik and {{\AA}}str{{\"o}}m, Karl Johan and Canudas de Wit, Carlos and G{{\"a}}fvert, Magnus and Lischinsky, Pablo},
  journal={European Journal of Control},
  volume={4},
  number={3},
  pages={176--195},
  year={1998},
  doi={10.1016/S0947-3580(98)70113-X}
}

\end{document}